\def\year{2017}\relax
\documentclass[letterpaper]{article}

\usepackage{aaai17}
\usepackage{times}
\usepackage{helvet}
\usepackage{courier}

\usepackage[pdftex]{graphicx}
\usepackage[cmex10]{amsmath}
\usepackage{algorithmic}

\usepackage[tight,footnotesize]{subfigure}
\usepackage{mathrsfs}
\usepackage{amssymb}
\usepackage{booktabs}

\usepackage{color}
\usepackage{threeparttable}
\usepackage{array}
\usepackage{url}
\usepackage{soul}

\usepackage{multirow}
\usepackage{tabularx}
\usepackage{tabulary}
\usepackage{dcolumn}

\begin{document}
 	

\title{An Empirical Analysis of Constrained Support Vector Quantile Regression\\ for Nonparametric Probabilistic Forecasting of Wind Power}
\author{}

\author{Kostas Hatalis, Shalinee Kishore\\
	Dept. of Electrical and	Computer Engineering\\
	Lehigh University\\
	\{kmh511,shk2\}@lehigh.edu\\
	\And
	Katya Scheinberg\\
	Dept. of Industrial Engineering\\
	Lehigh University\\
	kas410@lehigh.edu\\
	\And
	Alberto Lamadrid\\
	Dept. of Economics\\
	Lehigh University\\
	all512@lehigh.edu\\
}

\maketitle

\begin{abstract}
	Uncertainty analysis in the form of probabilistic forecasting can provide significant improvements in decision making processes in the smart power gird for better integrating renewable energies such as wind. Whereas point forecasting provides a single expected value, probabilistic forecasts provide more information in the form of quantiles, prediction intervals, or full predictive densities. This paper analyzes the effectiveness of an approach for nonparametric probabilistic forecasting of wind power that combines support vector machines and nonlinear quantile regression with non-crossing constraints. A numerical case study is conducted using publicly available wind data from the Global Energy Forecasting Competition 2014. Multiple quantiles are estimated to form 20\%, 40\%, 60\% and 80\% prediction intervals which are evaluated using the pinball loss function and reliability measures. Three benchmark models are used for comparison where results demonstrate the proposed approach leads to significantly better performance while preventing the problem of overlapping quantile estimates.
\end{abstract}

\section{Introduction}

Predicting and managing uncertainty in the production of wind power is one of the biggest challenges facing its integration into the smart grid. Forecasting uncertainty in wind is needed for many operational applications in a wind farm from turbine and storage control to bidding and trading in energy markets. Forecasting horizons can be categorized into three main time scales: short-term looking out several hours or days, long-term looking out to weeks or a month, and seasonal.  Traditionally wind power prediction is based on deterministic point forecasts where they provide an expected output for a given look-ahead time. These forecasts however lack uncertainty information. As such a large research effort has been taken recently by the renewables forecasting community \cite{hong2016probabilistic} to produce full probabilistic predictions which derive quantitative information on the associated uncertainty of power output. Although various methods have been proposed, it is still a challenge to make accurate and robust probabilistic predictions for highly nonlinear and complex data, such as wind.

Probabilistic wind models are based on either meteorological ensembles that are obtained by a weather model \cite{giebel2003using} or on statistical learning methods \cite{foley2012current}. Focusing on statistical learning, these methods can be applied to forecast full predictive distributions in the form of quantiles or prediction intervals. For instance, in \cite{pinson2004line} prediction intervals are estimated by adaptive re-sampling which is a common probabilistic forecasting strategy. Quantile regression (QR) is another very popular approach. In \cite{bremnes2004probabilistic} local QR is applied to estimate different quantiles while In \cite{nielsen2006using} spline based QR is used to estimate quantiles of wind power. In \cite{landry2016probabilistic} quantile loss gradient boosted machines are used to estimate 99 quantiles and in \cite{juban2016multiple} multiple quantile regression is used to predict a full distribution with optimization done using the alternating direction method of multipliers. 
A thorough overview of probabilistic wind power forecasting is provided in \cite{zhang2014review}.

In most of these approaches, estimation of each quantile is conducted independently. This could lead to the quantile cross over problem where a lower quantile overlaps a higher one. This is undesirable as it violates the principle of distribution functions where their associated inverse functions should be monotone increasing. A way to prevent this issue is to utilize a simple heuristic of reordering estimated quantiles, however this does not have much theoretical basis and may lead to inappropriate quantiles. 

The solution then is to optimize quantiles together with non-crossing constraints. In \cite{takeuchi2006nonparametric} a constrained support vector quantile regression (CSVQR) method was developed with non-crossing constraints where it was used to fit quantiles on static data. This formulation is re-purposed here for probabilistic forecasting. Other machine learning frameworks have been used before for uncertainty prediction of renewables such as nearest neighbors \cite{mangalova2016k}, neural networks \cite{sideratos2012probabilistic}, and extreme learning machines \cite{wan2014probabilistic} but support vector machines (SVMs) have yet to be examined for wind uncertainty forecasting. We propose that SVMs are not only effective in long term prediction due to their ability to handle nonlinear data via kernels but can be easily extended with constraints to ensure non-overlapping quantile estimates. Our study is the first to showcase the use of CSVQR with a sliding window of training data as well as showcase the effectiveness of constraints to ensure monotonically increasing quantiles for probabilistic prediction. We provide the derivation of CSVQR and analysis of experimental results on publicly available wind data. Several common benchmark methods are used for comparison. 

\section{Nonparametric Probabilistic Forecasting}

This sections highlights the underlying theory and evaluation methods used in probabilistic forecasting. For a random variable $Z_i$ such as wind power at time $ i $ its probability density function is defined as $f_i$ and its the cumulative distribution function as $F_i$. If $ F_{i} $ is a strictly increasing,  the quantile $ q_{\tau}(i) $ with proportion $ \tau \in [0,1] $ of the random variable $ Z_{i} $ is uniquely defined as the value $ x $ such that $ P(Z_i < x) = \tau $ or equivalently as the inverse of the distribution function $ q_{\tau}(i) = F_{i}^{-1}(\tau) $. A quantile forecast  $ \hat{q}_{\tau}(i+k) $ with nominal proportion $ \tau $ is an estimate of the true quantile $ q_{\tau}(i+k) $ for the lead time $ i+k $, given predictor values (such as numerical wind speed forecasts). Prediction intervals then give a range of possible values within which an observed value is expected to lie with a certain probability $ \beta \in [0,1] $. A prediction interval $ \hat{I}^{\beta}_{i+k} $ produced at time $ i $ for future horizon $ i +k $ is defined by its lower and upper bounds, which are the quantile forecasts $ \hat{I}^{\beta}_{i+k} = \left[ \hat{q}_{\tau_l}(i+k) , \hat{q}_{\tau_u}(i+k)\right]  $ whose nominal proportions $ \tau_l $ and $ \tau_u $ are such that $ \tau_u - \tau_l = 1-\beta $. 

If it is assumed the future density function will take a certain form then this is called parametric probabilistic forecasting. For a nonlinear and bounded process such as wind generation, probability distributions of future wind power for instance may be skewed and heavy-tailed distributions \cite{dorvlo2002estimating}. Else if no assumption is made about the shape of the distribution, a nonparametric probabilistic forecast $ \hat{f}_{i+k} $ \cite{pinson2007non} can be made of the density function by gathering a set of $ M $ quantiles forecasts such that $ \hat{f}_{i+k} = \left\lbrace \hat{q}_{m}(i+k),m=1,...,M|0\leq \tau_1 <  ... < \tau_M \leq 1 \right\rbrace  $ with chosen nominal proportions spread on the unit interval. In this paper we consider nonparametric forecasting of wind power on the resolution of one hour (predicting outwards to a month worth of values). On a short time scale of an hour, the wind density may fluctuate therefore making nonparametric forecasting more ideal then fitting a parametric density \cite{zhang2014review}.

\begin{figure}
	\centering
	\includegraphics[scale=0.5]{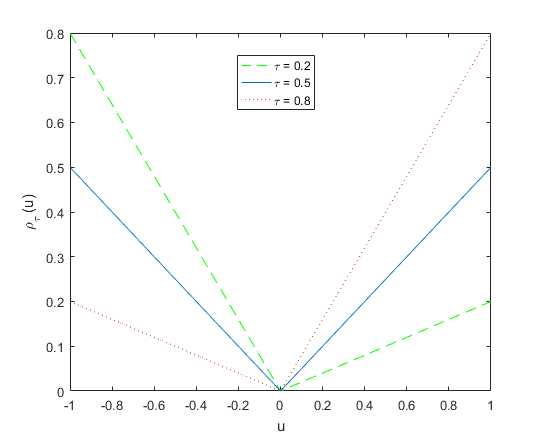}
	\caption{Plot of the pinball function for different $ \tau $ values.}
	\label{fig:pin}
\end{figure}

For nonparametric probabilistic forecasting quantile regression, introduced by \cite{koenker1978regression}, is a popular choice for estimating conditional quantiles. It is closely related to models for the conditional median \cite{koenker2005quantile}. Minimizing the mean absolute function leads to an estimate of the conditional median of a prediction. By applying asymmetric weights to errors through a tilted form of the absolute value function the conditional quantiles of a predictive distribution can be computed. To achieve this the pin ball loss function is used, which is defined by
\begin{equation*}
	\rho_{\tau}(u) = \left\lbrace 
	\begin{array}{cl}
		\tau u    & \mbox{if } u \geq 0 \\
		(\tau-1)u & \mbox{if } u < 0
	\end{array} \right.
\end{equation*}
\noindent where $ 0 < \tau < 1 $. A visualization of the pinball function with several different values of $ \tau $ is shown in Fig. \ref{fig:pin}. Given a vector of predictors $ {x}_i $ where $ i = 1,...,N $, weights $ {w} $ and
intercept $ b $ coefficient in a linear regression fashion, the conditional $ \tau $ quantile $ \hat{q}_\tau $ is given by	$ \hat{q}_\tau ({x}_i)= {w}^\top {x}_i +b $. The weights and intercept can be estimated by solving the following minimizing problem
\begin{equation} \label{eq3}
	\min \frac{1}{N} \sum_{i=1}^{N} \rho_\tau (y_i-\hat{q}_\tau ({x}_i))
\end{equation}
\noindent where $ y_i $ is the observed value of the predictand. The problem in Eq. (\ref{eq3}) can be minimized by linear programming. 

\subsection{Evaluation Methods}
In probabilistic forecasting it is important to evaluate the quantile estimates and derived predictive intervals. Prediction intervals (PIs) show where future wind power observations are expected to lie with an assigned probability termed as the PI nominal confidence (PINC) $ 100(1-\tau)\% $. The coverage probability of estimated PIs are expected to eventually reach a nominal level of confidence over the test data. A good measure for reliability which shows target coverage of the PIs is the PI coverage probability (PICP) which is defined by
\begin{equation*}
	PICP=\frac{1}{N}\sum_{i=1}^{N}c_i 
\text{ where }	
 c_i = \left\lbrace 
	\begin{array}{ll}
	1, & y_i \in I_{i}^{\beta}({x}_i)\\
	0, & y_i \notin I_{i}^{\beta}({x}_i)\\
	\end{array}\right.
\end{equation*}

\noindent  is the indicator of PICP and $N$ is the number of test samples. For reliable PIs, the examined PICP should be close to its corresponding PINC. A related assessment index is the average coverage error (ACE) which is defined by
\begin{equation*}
	ACE = |PICP-100\times(1-\beta)|
\end{equation*}
\noindent To ensure PIs with high reliability, the ACE should be as close to zero as possible. Next to evaluate quantile estimates and full predictive densities it is important to use the pinball function as an assessment score called the quantile score (Q-score). The Q-score is obtained for every estimated quantile and is averaged over all target quantiles for all future time steps. For a quantile forecast $ \hat{q}_\tau(t) $ the Q-score $ L_\tau (\hat{q}_\tau ,y) $ is defined as
\begin{equation*}
	L_\tau (\hat{q}_\tau ,y)=\left\lbrace 
	\begin{array}{ll}
		\frac{\tau}{100}(y-\hat{q}_\tau ) & \mbox{ if } y \geq \hat{q}_\tau \\
		\left( 1-\frac{\tau}{100}\right)(\hat{q}_\tau- y) & \mbox{ if } y < \hat{q}_\tau 	
	\end{array}
	\right. 
\end{equation*}

\noindent where $ y $ is the observation used for forecast evaluation. A lower Q-score indicates a better forecast. 

\section{Support Vector Quantile Regression}

To fit the nonlinearity of wind data, nonlinear quantile regression (NQR) can be utilized. NQR is implemented by projecting an input vector $ {x} $ into a potentially higher dimensional feature space $ \mathcal{F} $ using a nonlinear mapping function $ \phi(\cdot) $ implicitly defined by a kernel $ K $. This gives the functional form of $ f_{\tau}({x}) = w_{\tau}^{\top} \phi({x}) $ 
%
\noindent where $ f_{\tau} $ is the $ \tau $-th quantile of the distribution of $ y $ conditional on the values of $ {x} $, $ w_{\tau} $ is a vector of parameters. The NQR simplifies into linear quantile regression if  $ \phi({x})={x} $. To solve the NQR problem it can be expressed by the following formulation with added $ L_2 $ penalty to prevent overfitting
\begin{equation*}
	\min_{w_\tau} \frac{1}{2} \|w_{\tau}\|^2 + C \sum_{i=1}^{N} \rho_\tau (y_i-f_{\tau}(x_i))
\end{equation*}
\noindent By introducing slack variables $ \xi_i^{-}$ and $\xi_{i}^{+} $the problem can be re-written as a support vector quantile regression problem
\begin{equation} \label{eq5}
	\min_{w,b,\xi^{-},\xi^{+}} \frac{1}{2} \|w_\tau \|^2 + C \sum_{i=1}^{N} 
	(\tau \xi_{i}^{+} + (1-\tau) \xi_{i}^{-} )
\end{equation}
\begin{equation*}   
	\text{s.t. } \begin{cases} 
		y_i - w_{\tau}^{\top} \phi(x_i)-\xi_i^{+} \leq 0 \\ 
		-y_i + w_{\tau}^{\top} \phi(x_i)-\xi_i^{-}\leq 0 \\
		\xi_i^{-},\xi_i^{+}\geq 0
	\end{cases} 
\end{equation*}

\subsection{Non-crossing Quantile Constraints}

In Eq. (\ref{eq5}) a single quantile is estimated. To estimate multiple quantiles this formulation could be run to solve for different $ \tau $'s independently. However in doing so quantiles may cross each other which is not desirable since it violates the principle of monotone increasing inverse density functions. To prevent this, constraints need to be introduced \cite{takeuchi2006nonparametric}. $ 0< \tau_1<...<\tau_M $ are defined as the orders of $ M $ conditional quantiles to be estimated. To ensure these quantiles do not cross each other the following constraint is needed $ f_1(x_i) \leq ... \leq f_M(x_i), \forall_i $. With this constraint the primal problem of the non-crossing conditional quantile estimator is given by 
\begin{equation} \label{eqq}
	\min_{w,\xi^{-},\xi^{+}} 
	\sum_{m=1}^{M} \left( 
	\frac{1}{2} \|w_m \|^2 + C \sum_{i=1}^{N} 
	(\tau_m \xi_{mi}^{+} + (1-\tau_m) \xi_{mi}^{-} )
	\right) 
\end{equation}
\begin{equation*}    
	\text{s.t. } \begin{cases} 
		y_i - w_{m}^{\top} \phi(x_i)-\xi_{mi}^{+} \leq 0, & \forall_m, \forall_i \\ 
		-y_i + w_{m}^{\top} \phi(x_i)-\xi_{mi}^{-}\leq 0, & \forall_m, \forall_i \\
		\xi_{mi}^{-},\xi_{mi}^{+}\geq 0, & \forall_m, \forall_i \\
		w_{m}^{\top} \phi(x_i) - w_{m+1}^{\top} \phi(x_i) \leq 0, & \forall_m, \forall_i \\
	\end{cases} 
\end{equation*}

 
\noindent The Largrangian for the problem is then defined by
\begin{equation} \label{Lcross}\small
	\begin{split}
		L= &  \sum_{m=1}^{M}\left( \frac{1}{2} \|w_m \|^2 + C\sum_{i=1}^{N}(\tau_m \xi_{mi}^{+} + (1-\tau_m) \xi_{mi}^{-} )	\right. \\
		& + \sum_{i=1}^{N} \alpha_{mi}^{+}(y_i - w_{m}^{\top}\phi(x_i)-\xi_{mi}^{+})  \\
		& +\sum_{i=1}^{N} \alpha_{i}^{-}(-y_i + w_{m}^{\top} \phi(x_i)-\xi_{mi}^{-})\\
		& - \left. \sum_{i=1}^{N}(\eta_{mi}^{+}\xi_{mi}^{+}+\eta_{mi}^{-}\xi_{mi}^{-}) \right) \\
		&+  \sum_{m=1}^{M} \sum_{i=1}^{N} \lambda_{mi} \left( w_{m}^{\top} \phi(x_i) - w_{m+1}^{\top} \phi(x_i)  \right) \\
	\end{split}
\end{equation}

\noindent where a Lagrange multiplier $ \lambda_{mi}\geq 0 $ is introduced for $ m=1,...,M-1 $, $ \forall_i $, and  $ \lambda_{0i}=\lambda_{Mi} = 0 $. By letting the partial derivatives of $ L $ with respect to $ w_m $ be zero, the following is obtained
\begin{equation}
	\begin{split}
	& \frac{\partial L}{\partial w_m} =	w_m - \sum_{i=1}^{N} (\alpha_{mi}^{+}-\alpha_{mi}^{-})\phi(x_i) \\
	& + \sum_{i=1}^{N} (\lambda_{mi} -\lambda_{m-1i})\phi(x_i)=0
	\end{split}
\end{equation}

\noindent Partial derivatives of the other primal variables $ \xi_{mi}^{+} $ and $\xi_{mi}^{-} $ are 
\begin{align}
	\frac{\partial L}{\partial \xi_{mi}^{+}} &= \tau_m C - \alpha_{mi}^{+} - \eta_{mi}^{+} = 0 \label{xi1}\\
	\frac{\partial L}{\partial \xi_{mi}^{-}} &= (1-\tau_m) C - \alpha_{mi}^{-} - \eta_{mi}^{-} = 0 \label{xi2}
\end{align}

\noindent Plugging these equalities back into Eq. (\ref{Lcross}) the following dual minimization problem can be obtained

\begin{equation}\label{eqdualcross} \small
\begin{split}
	& \min_{\alpha^{+},\alpha^{-},\lambda} \sum_{m=1}^{M} \left( 
	-\frac{1}{2} \sum_{i=1}^{N} \sum_{j=1}^{N} (\alpha_{mi}^{+}-\alpha_{mi}^{-}) (\alpha_{mj}^{+}-\alpha_{mj}^{-})... \right.\\
	& K(x_i,x_j) + \sum_{i=1}^{N}(\alpha_{mi}^{+}-\alpha_{mi}^{-})y_i \\
	&  -\frac{1}{2} \sum_{i=1}^{N}\sum_{i=j}^{N} (\lambda_{mi} -\lambda_{m-1i})(\lambda_{mj} - \lambda_{m-1j})K(x_i,x_j) \\
	& +\left.
	\sum_{i=1}^{N}\sum_{i=j}^{N}(\alpha_{mi}^{+}-\alpha_{mi}^{-})(\lambda_{mj} - \lambda_{m-1j})K(x_i,x_j) \right) 
\end{split}
\end{equation}
\begin{equation*}    \small
	\text{subject to} \begin{cases}
		\lambda_{mi} \geq 0 , \forall_m \forall_i\\
		\alpha_{mi}^{+} \in [0,\tau_m C], \forall_m \forall_i \\ 
		\alpha_{mi}^{-} \in [0,(1-\tau_m) C], \forall_m \forall_i
	\end{cases}
\end{equation*}

\noindent From this dual formulation the conditional quantile $ \tau_m $ can then be given by
\begin{equation}
\begin{split}
	& f_{\tau_m}(x) =  \sum_{i=1}^{N} (\alpha_{mi}^{+}-\alpha_{mi}^{-})K(x,x_i) \\
	& - \sum_{i=1}^{N} (\lambda_{mi} -\lambda_{m-1i})K(x,x_i)
	\end{split}
\end{equation}

\noindent Since the dual form is a quadratic programming (QP) problem it can be solved by a number of QP methods. For testing the constrained SVQR (CSVQR) method the radial basis function (RBF) kernel is utilized as it is a popular kernel function choice for support vector machines. Other kernels were tested on the case data sets described in the next section but resulted in poor results. The RBF kernel, given two samples $ \mathbf{x}  $ and $ \mathbf{x'} $ which are represented as feature vectors, is calculated as
\begin{equation*}
K(\mathbf{x}, \mathbf{x'}) =  \phi(\mathbf{x})^\top \phi(\mathbf{x'})=
\exp \left(-{\frac {||\mathbf {x} -\mathbf {x'} ||^{2}}{2\sigma ^{2}}}\right) 
\end{equation*}

\noindent An advantage of a RBF kernel is that it can project vectors into an infinite dimensional feature space. In order to quickly solve for conditional quantile estimates sequential minimization optimization \cite{platt1998sequential} is applied to Eq. (\ref{eqdualcross}).

%

%

\section{Application To The GEFCom2014 Dataset}

\begin{figure}
	\centering
	\includegraphics[scale=0.25]{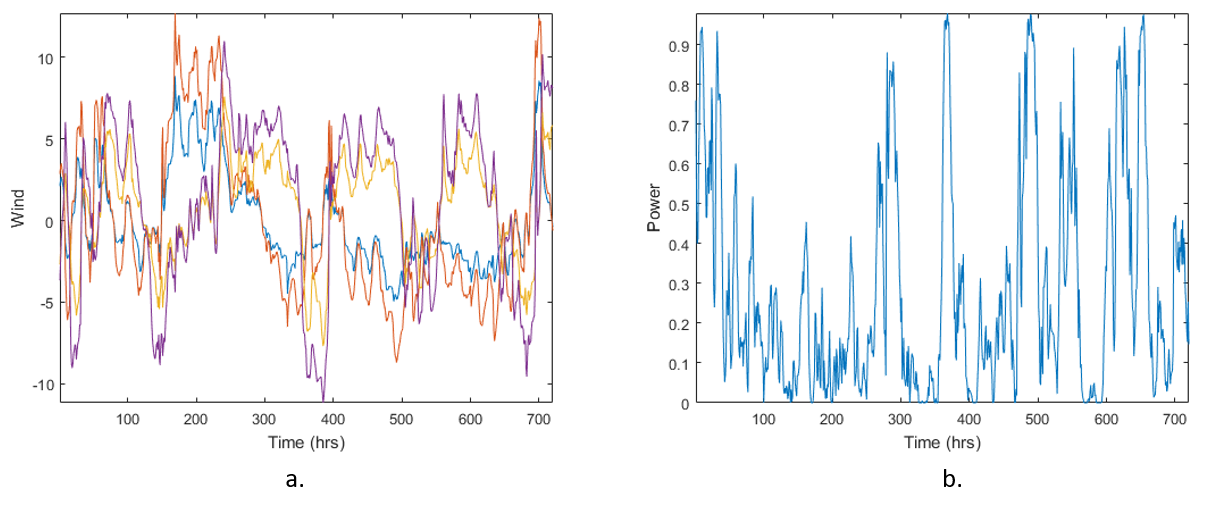}
	\caption{(a) Example plot of numerical wind predictions at 10m and 100m for U and V directions used as inputs to forecast wind power. (b) Observed wind power corresponding to the same time stamps.}
	\label{fig:gef}
\end{figure}

Data for this case study comes from the publicly available Global Energy Forecasting Competition 2014 \cite{hong2016probabilistic}. The goal of the competition was to design parametric or nonparametric forecasting methods that would allow conditional predictive densities of the wind power generation to be described as a function of input data which were future weather forecasts and/or past wind power. Data is provided for the years of 2012 and 2013 from 10 wind farms titled Zone 1 to Zone 10. The predictors are numerical weather predictions (NWPs) in the form of wind speeds at an hourly resolution at two heights, 10m and 100m above ground level. These forecasts are for the zonal and meridional wind components (denoted U and V). It was up to users to deduce exact wind speed, direction, and other wind features if necessary. These NWPs were provided for the exact locations of the wind farms. Additionally, power measurements at the various wind farms, with an hourly resolution, are also provided. All power measurements are normalized by the nominal capacity of their wind farm. The goal in forecasting was to learn to associate the provided NWPs (or derived features) with wind power. Then NWPs are provided for the forecasting horizon of one month and it is up to a learning model to use those NWPs as input to a learning model to predict quantiles at each future time step. Fig. \ref{fig:gef} showcases an example month worth of data where Fig. \ref{fig:gef}.a shows the four NWP given and Fig. \ref{fig:gef}.b shows their corresponding normalized wind power output.

In our analysis of CSVQR we used the summer months of June 2013 to August 2013 and fall months of September 2013 to November 2013 for testing from Zone 1. Training was done using a sliding window of three previous months to forecast the fourth month. For instance to predict June training was done on observed data from March to May, then to predict July training was done from April to June, etc. Thirteen features were derived from the raw data for training the CSVQR model. Features used are derived wind speeds at 10m and 100m, wind direction at 10m and 100m, wind energy at 10m and 100m, wind shear, wind energy difference (between 10m and 100m), wind direction difference (between 10m and 100m), and  included in training are also the four raw wind speeds at 10m and 100m for U and V directions. All features were normalized between 0 and 1. Denoting $ u $ and $ v $ as the wind components and $ d $ as the energy density (we used $ d=1 $), the equations used to compute wind speed (ws), wind direction (wd), wind energy (we), and wind shear (wsh) are
\begin{equation*}
\begin{split}
\text{ws} & =  \sqrt{u^2 + v^2}\\
\text{wd} & = \frac{180}{\pi} \times \arctan(u,v)\\
\text{we} & = \frac{1}{2} \times d \times \text{ws}^3\\
\text{wsh} & = \sqrt{\text{ws10}^2 + \text{ws100}^2}
\end{split}
\end{equation*}

To empirically analyze the CSVQR model as an appropriate method for wind forecasting it is compared with two industry models and a naive model that are used for benchmarking in probabilistic wind forecasting applications \cite{sideratos2012probabilistic,pinson2007non,pinson2010conditional}. The first is called the persistence method which is the most common benchmark and is considered difficult to outperform for short-term forecasting. This method corresponds to the persistence distribution and is formed by the most recent observations. For this case study, the past 12 hours of wind power observations were used to form the persistence distribution. Second method is the climatology approach where its predictive distribution is unconditional and based on all available past wind power observations. It is considered harder to beat in long-term forecasting. Lastly, the uniform distribution is used for a naive benchmark method where it assumes all wind power values at each time step occur with equal probability.

\begin{figure*}
	\centering
	\includegraphics[scale=0.45 ]{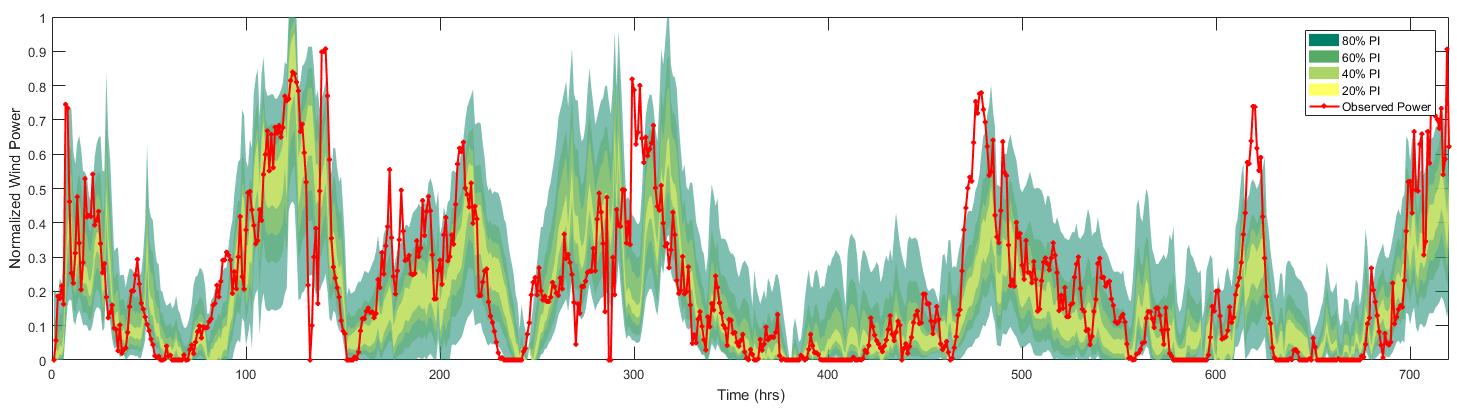}
	\caption{Example plot of estimated 80\%, 60\%, 40\%, and 20\% prediction intervals along with observed wind power in red for the month of July 2013.}
	\label{fig:fanchart}
\end{figure*}

\subsection{Results}

To visualize a probabilistic forecast Fig. \ref{fig:fanchart} shows an example prediction for 80\%, 60\% 40\%, and 20\% prediction intervals for the month of July 2013. Observed wind power is shown in red. From such probabilistic forecasts it is then possible to derive full predictive density functions following that the estimated conditional quantiles are nondecreasing  \cite{quinonero2006evaluating}. 
Evaluation results for reliability of probabilistic forecasts in the form of prediction intervals of wind power over the months of June 2013 to November 2013 is shown in Table 1. Results are shown for the CSVQR method and for the climatology, persistence, and uniform benchmark methods. Evaluation metrics for the PINC are the PICP and ACE. For the month of June and October, the climatology method was slightly better but this was due to the fact that this model can yield wide intervals to cover more data. However in all other months CSVQR outperformed all three benchmarks by several magnitudes. To further fully evaluate the forecasts it is also important to look at the quantile score to measure the coverage of the estimated quantiles. Table 2 shows the summary of Q-scores averaged across all quantiles from all lookahead periods for every forecast month. Their standard deviation is also provided to quantify the amount of variation among the quantiles. The Q-scores of the proposed approach was very low and gave excellent probabilistic forecasts across all different months.

\section{Discussion}

Wind power forecasting is crucial for many decision making problems in power systems operations, and is a vital component in integrating more wind into the power grid. Due to the chaotic nature of the wind it is often difficult to forecast. Uncertainty analysis in the form of probabilistic wind prediction can provide a better picture of future wind coverage. This paper studies a framework for probabilistic forecasting using support vector quantile regression with non-crossing constraints to ensure multiple quantiles can be predicted without overlapping each other. Effectiveness of the CSVQR approach is validated with the real world dataset of the Global Energy Forecasting Competition 2014. Forecasts are compared to common benchmarks and are evaluated using the quantile score and reliability metrics. Results show adequate reliability and low quantile scores across the prediction horizon, which verify effectiveness of the model for forecasting while preventing estimated quantiles from overlapping. Furthermore, this approach has the potential to be applied across a variety of domains. Future work will look into applying CSVQR to forecast electricity pricing and load demand for smart grid applications.

\renewcommand{\arraystretch}{0.6}
\begin{table*}[h]
	\centering
	\begin{tabular*}{\textwidth}{@{\extracolsep{\fill}}rrrrrrrrrr@{}}
		\toprule
		\multirow{3}{*}{Month} & \multirow{3}{*}{PINC} & \multicolumn{2}{c}{CSVQR} & \multicolumn{2}{c}{Climatology} & \multicolumn{2}{c}{Persistence} & \multicolumn{2}{c}{Uniform}  \\
		\cline{3-4} \cline{5-6} \cline{7-8} \cline{9-10} & & & & & & & & & \\
		& & PICP & ACE & PICP & ACE & PICP & ACE & PICP & ACE\\
		
		\midrule
		\multirow{4}{*}{June 13} & 80\% & \textbf{85.00} & \textbf{5.00} & 95.28 & 15.28 & 46.11 & 33.89 & 60.97 &  19.03\\
		& 60\% & 66.25 & 6.25 & \textbf{62.50} & \textbf{2.50} & 37.64 & 22.36 & 40.97 & 19.03 \\
		& 40\% & 45.56 & 5.56 & \textbf{42.92} & \textbf{2.92} & 30.56 & 9.44 & 23.47 & 16.53 \\
		& 20\% & 25.42 & 5.42 & \textbf{22.64} & \textbf{2.64} & 26.30 & 6.31 & 10.69 & 9.31 \\
		
		\midrule
		\multirow{4}{*}{July 13} & 80\% & \textbf{78.49} & \textbf{1.50} & 76.08 & 3.92 & 12.77 & 67.23  & 59.27 & 20.73 \\
		& 60\% & \textbf{56.04} & \textbf{3.95} & 55.38 & 4.62 & 6.72 & 53.28 & 36.96 & 23.04 \\
		& 40\% & \textbf{38.70} & \textbf{1.29} & 35.08 & 4.92 & 5.24 & 34.76 & 21.91 & 18.09 \\
		& 20\% & \textbf{20.96} & \textbf{0.96} & 16.80 & 3.20 & 2.55 & 17.45 & 10.08 & 9.92 \\
		
		\midrule
		\multirow{4}{*}{August 13} & 80\% & \textbf{78.36} & \textbf{1.64} & 65.73 & 14.27 & 22.04 & 57.96 & 61.83 & 18.17 \\
		& 60\% & \textbf{59.27} & \textbf{0.73} & 42.61 & 17.39 & 13.44 & 46.56 & 44.49 & 15.51 \\
		& 40\% & \textbf{40.46} & \textbf{0.46} & 25.94 & 14.06 & 7.80  & 32.20 & 30.11 & 9.89 \\
		& 20\% & \textbf{19.89} & \textbf{0.11} & 9.95  & 10.05 & 4.57  & 15.43 & 15.05 & 4.95 \\
		
		\midrule
		\multirow{4}{*}{September 13} & 80\% & \textbf{79.03} & \textbf{0.97} & 81.81 & 1.81 & 31.53 & 48.47 & 60.69 & 19.31 \\
		& 60\% & \textbf{60.69} & \textbf{0.69} & 59.30 & 0.70 & 23.75 & 36.25 & 35.56 & 24.44 \\
		& 40\% & \textbf{42.92} & \textbf{2.92} & 34.31 & 5.69 & 14.86 & 25.14 & 20.97 & 19.03 \\
		& 20\% & \textbf{22.36} & \textbf{2.36} & 15.83 & 4.17 & 5.97 & 14.03 & 9.31 & 10.69 \\
		
		\midrule
		\multirow{4}{*}{October 13} & 80\% & 83.20 & 3.20 & \textbf{81.85} & \textbf{1.85} & 52.82 & 27.18 & 62.77 & 17.23 \\
		& 60\% & 68.15 & 8.15 & \textbf{62.77} & \textbf{2.77} & 23.92 & 36.08 & 45.70 & 14.30 \\
		& 40\% & 52.55 & 12.55 & \textbf{46.24} & \textbf{6.24} & 6.85 & 33.15 & 28.76 & 11.24 \\
		& 20\% & \textbf{24.36} & \textbf{4.36} & 25.27 & 5.27 & 1.88 & 18.12 & 16.67 & 3.33 \\
		
		\midrule
		\multirow{4}{*}{November 13} & 80\% & \textbf{80.42} & \textbf{0.42} & 90.14 & 10.14 & 25.83 & 54.17 & 72.36 & 7.64 \\
		& 60\% & \textbf{59.31} & \textbf{0.69} & 75.00 & 15.00 & 15.14 & 44.86 & 48.75 & 11.25 \\
		& 40\% & \textbf{36.11} & \textbf{3.89} & 55.69 & 15.69 & 11.94 & 28.06 & 29.17 & 10.83 \\
		& 20\% & \textbf{16.53} & \textbf{3.47} & 29.03 & 9.03  & 10.42 & 9.58 & 13.19 & 6.81 \\
		\bottomrule
	\end{tabular*}
	\caption{Results of prediction interval reliability in different months.}
\end{table*}

\begin{table}[h] 
	\centering
	\begin{tabular}{cccc}
		\toprule
		Month & Method & Q-Score & SD \\
		\midrule
		\multirow{3}{*}{June 13} & CSVQR 		& \textbf{0.0404} & \textbf{0.0119} \\
		& Climatology  & 0.0628 & 0.0230 \\
		& Persistence	& 0.0880 & 0.0406 \\
		& Uniform 		& 0.1105 & 0.0434 \\
		\midrule
		\multirow{3}{*}{July 13} & CSVQR 		& \textbf{0.0546} & \textbf{0.0169} \\
		& Climatology 	& 0.1038 & 0.0401 \\
		& Persistence 	& 0.1799 & 0.0681 \\
		& Uniform 		& 0.1112 & 0.0428 \\
		\midrule
		\multirow{3}{*}{August 13}  & CSVQR 		& \textbf{0.0677} & \textbf{0.0199} \\
		& Climatology 	& 0.1374 & 0.0555 \\
		& Persistence 	& 0.1734 & 0.0738 \\
		& Uniform 		& 0.1033 & 0.0380 \\
		\midrule
		\multirow{3}{*}{September 13} & CSVQR 		& \textbf{0.0590} & \textbf{0.0172} \\
		& Climatology & 0.0992 & 0.0401 \\
		& Persistence & 0.1659 & 0.0582 \\
		& Uniform 	& 0.1107 & 0.0429 \\
		\midrule
		\multirow{3}{*}{October 13}     & CSVQR 		& \textbf{0.0561} & \textbf{0.0159} \\
		& Climatology 	& 0.0971 & 0.0366 \\
		& Persistence 	& 0.1807 & 0.0977 \\
		& Uniform 		& 0.1033 & 0.0382 \\
		\midrule
		\multirow{3}{*}{November 13}  & CSVQR 		& \textbf{0.0557} & \textbf{0.0186} \\
		& Climatology & 0.0844 & 0.0396 \\
		& Persistence & 0.1089 & 0.0533 \\
		& Uniform 	& 0.0978 & 0.0406 \\
		\midrule
		\multirow{3}{*}{All}  		  & CSVQR 		& \textbf{0.0556} & \textbf{0.0167} \\
		& Climatology & 0.0974 & 0.0391 \\
		& Persistence & 0.1494 & 0.1261 \\
		& Uniform 	& 0.1061 & 0.0409 \\			      
		\bottomrule
	\end{tabular}
	\caption{Summary of the mean Q-score across all quantiles for a given method and month and their standard deviation.}
\end{table}

\bibliographystyle{aaai}
\bibliography{mybib}

\end{document}